\documentclass{bmvc2k}
\usepackage{amsmath}
\usepackage{amssymb}
\usepackage{booktabs}
\usepackage{multirow}
\usepackage{placeins}

\title{MultiFlow: Vision-Driven Multimodal fMRI Encoding via Stimulus-Informed Source Flow Matching}

\addauthor{Bui Van Hung}{hungbv16@jnu.ac.kr}{1}
\addauthor{Hoang-Son Vo}{hoangsonvothanh@jnu.ac.kr}{1}
\addauthor{Mai Duc Minh Huy}{minhhuymaiduc@gmail.com}{2}
\addauthor{Dang Thanh Vu}{dtvu@aiseed.kr}{3}
\addauthor{Soo-Hyung Kim}{shkim@jnu.ac.kr}{1}

\addinstitution{
 Chonnam National University\\
 Gwangju, South Korea
}
\addinstitution{
 Ho Chi Minh University of Science\\
 Ho Chi Minh City, Vietnam
}
\addinstitution{
 AISEED, Inc.\\
 South Korea
}

\runninghead{Bui et al.}{MultiFlow: Stimulus-Informed Source Flow Matching}

\begin{document}

\maketitle
\begin{center}\small\emph{Preprint. Accepted at the British Machine Vision Conference (BMVC) 2026.}\end{center}

\begin{abstract}
Current fMRI encoding models frame prediction as deterministic regression, yet identical stimuli evoke a distribution of neural responses (arising from physiological noise, hemodynamic variability, and cognitive fluctuations) that a single point estimate cannot capture. This work introduces \textbf{MultiFlow}, a generative fMRI encoding framework based on \textbf{Conditional Flow Matching (CFM)} and a formulation termed \textbf{Stimulus-Informed Source Flow Matching (SISFM)}. SISFM generalises CFM by conditioning the source distribution on the stimulus: a learnable temporal convolution shapes the source mean with an HRF-length receptive field, while a variance head shrinks its scale, shortening the expected transport path. SISFM is instantiated with a Diffusion Transformer (DiT-X) decoder parameterising the velocity field, with sub-layer Attention Residuals, heterogeneous fusion of ten frozen vision, audio, language, and audio-visual backbones, and per-subject Yeo-network heads. On the Algonauts 2025 dataset, a single MultiFlow checkpoint recovers most of the benefit of large deterministic ensembles: it trails the 1000-model TRIBE ensemble in-distribution by $0.003$ PCC ($0.3163$ vs.\ $0.3195$), and the strongest in- and out-of-distribution ensembles (VIBE-20, MedARC-50) by $0.010$ and $0.009$ PCC, using a single trained model rather than an ensemble of 20--1000. The same checkpoint also provides stimulus-conditioned uncertainty estimates and sample-based diversity.
\end{abstract}

\section{Introduction}
\label{sec:intro}

\begin{figure}[!htbp]
\centering
\includegraphics[width=0.72\linewidth]{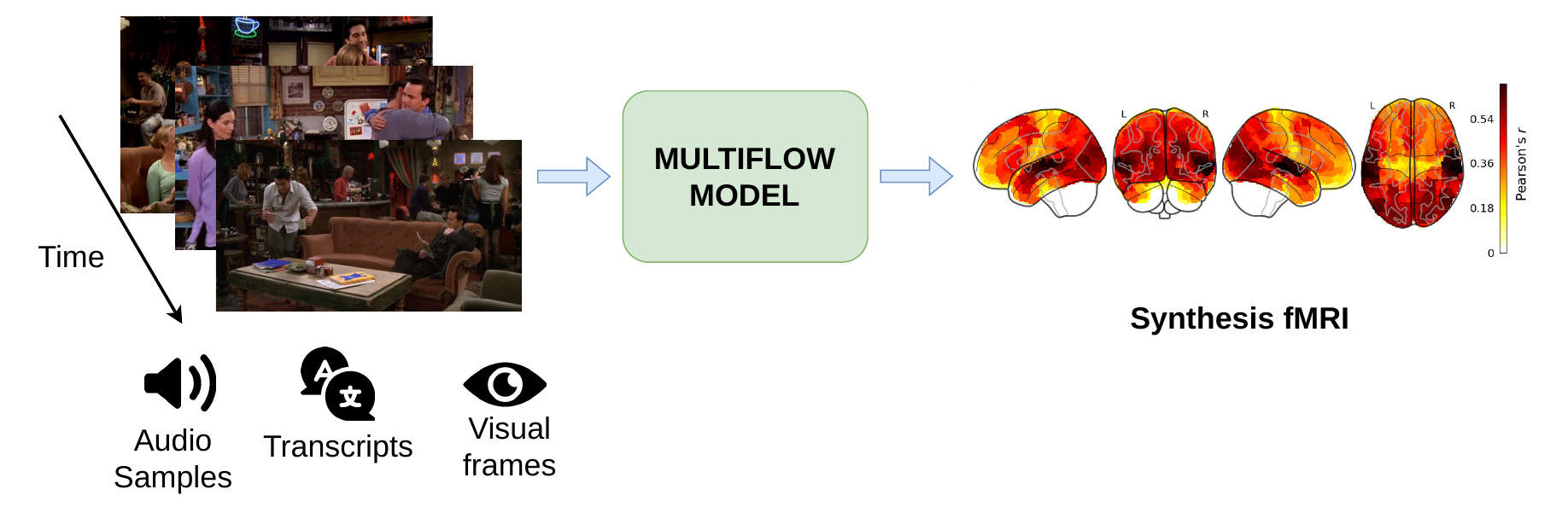}
\caption{\textbf{Overview of the MultiFlow pipeline.}
Time-aligned \emph{visual frames}, \emph{audio samples}, and \emph{transcripts} pass through ten frozen foundation-model backbones; the fused representation drives a generative model that synthesises whole-brain fMRI activity over the Schaefer-1000 parcellation. Rather than a single point estimate, MultiFlow models the stimulus-conditioned BOLD distribution via Stimulus-Informed Source Flow Matching (SISFM).}
\label{fig:overview}
\end{figure}

Naturalistic paradigms in which participants watch movies during fMRI scanning have become a standard setting for studying how the brain integrates vision, language, and sound over time (Figure~\ref{fig:overview}). Encoding models that predict cortical responses from such stimuli give a principled way to probe this integration~\cite{Yamins2016}, and the Algonauts 2025 dataset~\cite{Algonauts2025,CNeuroMod}, with tens of hours of whole-brain recordings per participant, makes it feasible to train deep encoders on them.

Concretely, the task is the following. A participant lies in an MRI scanner watching an episode of a television series. Every $1.49$\,s the scanner records one whole-brain volume, summarised here as $1000$ cortical parcels. An \emph{encoding model} sees only what the participant saw and heard, that is the video frames, the soundtrack and the dialogue, and must predict what the scanner recorded, parcel by parcel and timepoint by timepoint; accuracy is the correlation between predicted and measured signal in each parcel. A model that does this well is a testable account of how sensory input drives cortical activity, and lets stimuli or hypotheses be screened without acquiring new scans.

Leading methods on this dataset (TRIBE~\cite{TRIBE}, VIBE~\cite{VIBE} and MedARC~\cite{MedARC}) establish multimodal fusion as essential for high-fidelity encoding, yet all treat encoding as deterministic regression to a single BOLD point estimate. This discards a defining property of cortical responses: identical stimuli evoke a \emph{distribution} of neural responses. Squared-error optimisation collapses that distribution onto the conditional mean, a collapse hypothesised to underlie the limited out-of-distribution (OOD) generalisation of current encoders~\cite{AlgonautsInsights}.

\begin{figure}[!htbp]
\centering
\includegraphics[width=0.68\linewidth]{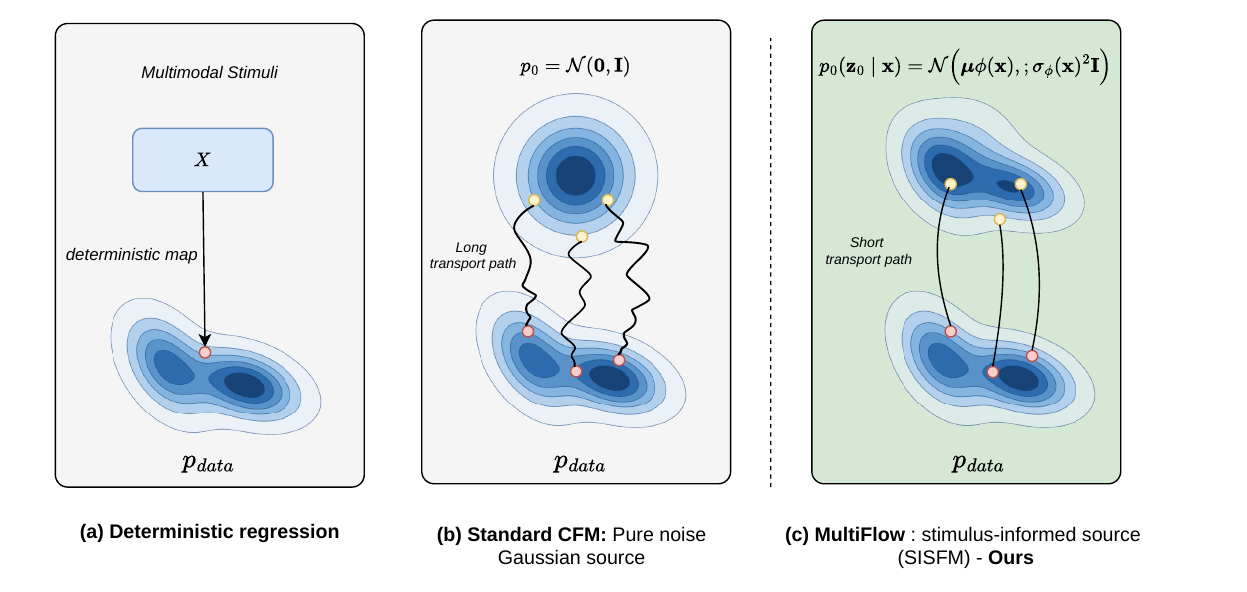}
\caption{\textbf{Three views of fMRI encoding (schematic).}
\textbf{(a)} Deterministic regression maps the stimulus $\mathbf{x}$ to a single
point estimate $\hat{\mathbf{y}}$ (red), collapsing the target BOLD distribution
onto its conditional mean.
\textbf{(b)} Standard CFM draws source samples $\mathbf{z}_0$ (orange) from a fixed
Gaussian $p_0=\mathcal{N}(\mathbf{0},\mathbf{I})$ and transports each to
$p_\text{data}\!\approx\!p(\mathbf{y}\mid\mathbf{x})$ along a long ODE path.
\textbf{(c)} \textbf{MultiFlow (ours)} draws the \emph{same} samples from a
\emph{stimulus-conditioned} source
$p_0(\mathbf{z}_0\mid\mathbf{x})=\mathcal{N}\!\bigl(\boldsymbol{\mu}_\phi(\mathbf{x}),\sigma_\phi(\mathbf{x})^2\mathbf{I}\bigr)$
whose $\boldsymbol{\mu}_\phi$ and $\sigma_\phi$ are \emph{learned} to place the
source closer to the target, shortening the expected transport path. In (b,c) the
curves are independent samples from a \emph{single} model; the panels differ only in
the source distribution.}
\label{fig:teaser}
\end{figure}

Encoding is therefore recast as a \emph{conditional generative} problem using flow matching~\cite{Lipman2022,Liu2022,Tong2023} (Figure~\ref{fig:teaser}). Standard CFM inherits a Gaussian source $p_0=\mathcal{N}(\mathbf{0},\mathbf{I})$ from image synthesis, where no stimulus exists at inference; in encoding the stimulus is observed at \emph{both} training and inference, yet a Gaussian source ignores it. This work introduces \textbf{Stimulus-Informed Source Flow Matching (SISFM)}, which conditions the source on the stimulus: a depthwise temporal convolution shapes the source mean with a receptive field matching the hemodynamic response function (HRF) length, while a Softplus variance head shrinks its scale, shortening the expected transport path. SISFM subsumes vanilla CFM as the special case $\boldsymbol{\mu}_\phi\!\equiv\!\mathbf{0}$, $\sigma_\phi\!\equiv\!1$.

SISFM is instantiated as \textbf{MultiFlow}: a heterogeneous fusion module over ten frozen vision, audio, language, and audio-visual foundation models~\cite{vJEPA2,DINOv2,InternVL3,VLM2Vec,Whisper,Llama3,Qwen25,Qwen25Omni}, a DiT-X decoder~\cite{ManiFlow,Peebles2023} parameterising the velocity field with sub-layer Attention Residuals~\cite{AttentionResiduals} in place of fixed identity skips, and per-subject output heads partitioned by the seven Yeo functional networks~\cite{Yeo2011} of the Schaefer-1000 parcellation~\cite{Schaefer2018}.

The primary contributions are:
\begin{itemize}
\item \textit{SISFM}: a generalisation of CFM that conditions the source
  distribution on the stimulus through an HRF-length depthwise convolution and a
  Softplus variance head, trained jointly with the velocity field via a
  three-term objective; to our knowledge, this is the first flow-matching framework
  for fMRI encoding.
\item The \textit{MultiFlow architecture}: a DiT-X decoder with sub-layer
  Attention Residuals~\cite{AttentionResiduals}, heterogeneous multi-token fusion
  over ten frozen backbones, and per-subject Yeo-network heads.
\item An \textit{empirical evaluation} on Algonauts 2025: a single MultiFlow
  checkpoint trails the 1000-model TRIBE ensemble in-distribution by $0.003$ PCC
  and the strongest OOD ensemble by $0.009$ PCC without model averaging and
  without claiming parity (Tables~\ref{tab:indist}--\ref{tab:ood}), and also
  delivers stimulus-conditioned uncertainty and sample-based diversity.
\end{itemize}

\section{Related Work}
\label{sec:related}

\noindent\textbf{fMRI encoding from naturalistic stimuli.} 
Classical encoding models map stimulus features to voxel-wise BOLD responses via ridge regression~\cite{Huth2012,Huth2016,Yamins2016}. The three leading entries on Algonauts 2025 share a common recipe: extract time-aligned features from frozen video, audio, and language foundation models, fuse them, and regress the result onto parcel responses with a transformer. They differ in fusion, readout, and ensembling. TRIBE~\cite{TRIBE} mean-pools per-modality features before fusion, reads out through a single shared per-subject linear head, and averages $1000$ runs. VIBE~\cite{VIBE} uses a similar frozen-feature pipeline and averages $20$ runs; both rely on hard gradient-stopping to keep auxiliary objectives from dominating the encoder. MedARC~\cite{MedARC} averages $50$ runs and is the strongest published OOD ensemble. All three treat encoding as deterministic regression to a single BOLD point estimate. This discards BOLD variability arising from physiological noise~\cite{Friston2003}, hemodynamic dispersion~\cite{Glover1999}, and cognitive variability, limiting out-of-distribution (OOD) generalisation~\cite{AlgonautsInsights}. MultiFlow recasts encoding as a conditional generative task to model this full response distribution.

\noindent\textbf{Generative models of neural data.} 
Generative models (e.g., diffusion) are widely used for fMRI \emph{decoding} (brain-to-stimulus)~\cite{BrainDiffuser,MindEye,MindEye2}. Elsewhere in medical imaging a generative branch often serves as an auxiliary regulariser rather than as the primary output, as in the VAE-regularised CNN-Transformer of SegTransVAE~\cite{SegTransVAE}; we observe a related effect from the SISFM objective (Section~\ref{sec:uq}). Encoding (stimulus-to-brain), however, remains under-explored generatively. MultiFlow is a conditional flow-matching (CFM) encoder, modelling the stimulus-conditioned BOLD distribution $p(\mathbf{y}\mid\mathbf{x})$.

\noindent\textbf{Conditional Flow Matching (CFM) and SISFM.} 
CFM~\cite{Lipman2022,Liu2022,Tong2023} learns a velocity field $v_\theta(\mathbf{z}_t, t, \mathbf{x})$ that generates data $\mathbf{y}$ from a source $\mathbf{z}_0 \sim p_0$ via the ODE $\dot{\mathbf{z}}_t = v_\theta$. Under the optimal-transport coupling, the paths are defined by the linear interpolation:
\begin{equation}
\mathbf{z}_t = t\,\mathbf{y} + (1-t)\,\mathbf{z}_0,\qquad u_t = \mathbf{y}-\mathbf{z}_0.
\label{eq:cfm-path}
\end{equation}
The model is trained by regressing $v_\theta$ onto the target velocity $u_t$:
\begin{equation}
\label{eq:cfm-loss}
\mathcal{L}_\text{flow}(\theta;\phi) = \mathbb{E}_{t,\mathbf{x},\mathbf{y},\mathbf{z}_0\sim p_0(\cdot\mid\mathbf{x})}\,\bigl\|v_\theta(\mathbf{z}_t,t,\mathbf{x})-(\mathbf{y}-\mathbf{z}_0)\bigr\|^2.
\end{equation}
Standard CFM fixes the source to a stimulus-blind Gaussian $p_0 = \mathcal{N}(\mathbf{0}, \mathbf{I})$, which in encoding ignores an input that is available at inference, increasing the expected transport distance $\mathbb{E}\,\|\mathbf{y}-\mathbf{z}_0\|^2$ and causing trajectories to cross. We introduce \textbf{Stimulus-Informed Source Flow Matching (SISFM)}, which conditions the source on the stimulus:
\begin{equation}
p_0(\mathbf{z}_0\mid\mathbf{x}) = \mathcal{N}\!\bigl(\boldsymbol{\mu}_\phi(\mathbf{x}),\,\sigma_\phi(\mathbf{x})^2\,\mathbf{I}\bigr),
\label{eq:sisfm-source}
\end{equation}
where the mean $\boldsymbol{\mu}_\phi$ and variance scale $\sigma_\phi$ are learnable: $\boldsymbol{\mu}_\phi$ acts as a shape prior and $\sigma_\phi$ shrinks the source ball, leaving the velocity field to model residual variability. Section~\ref{sec:sisfm} gives the parameterisation.

\FloatBarrier
\section{Method}
\label{sec:method}

\subsection{Problem Formulation and Notation}
We now state the task formally. A subject watches a movie sampled at TR $\approx 1.49$\,s. For each window, the input is a multimodal stimulus context $\mathbf{x}\in\mathbb{R}^{T_\text{ctx}\times D}$ ($T_\text{ctx}=101$ TRs, $D=31{,}872$ features from ten frozen backbones). The target is the corresponding cortical response $\mathbf{y}\in\mathbb{R}^{T_\text{tgt}\times V}$ ($T_\text{tgt}=50$ TRs, $V=1000$ Schaefer-1000 parcels). The context window is centred on the target with symmetric padding ($\approx 37$\,s on each side) to provide narrative history and brief look-ahead. Rather than using deterministic regression to predict a single BOLD point estimate $\hat{\mathbf{y}}=f_\theta(\mathbf{x})$, MultiFlow models the BOLD distribution $p(\mathbf{y}\mid\mathbf{x})$ via flow matching, starting from the stimulus-conditioned source distribution in Eq.~\eqref{eq:sisfm-source}.

\begin{figure*}[t]
\centering
\includegraphics[trim=20bp 13bp 208bp 13bp, clip, width=\linewidth]{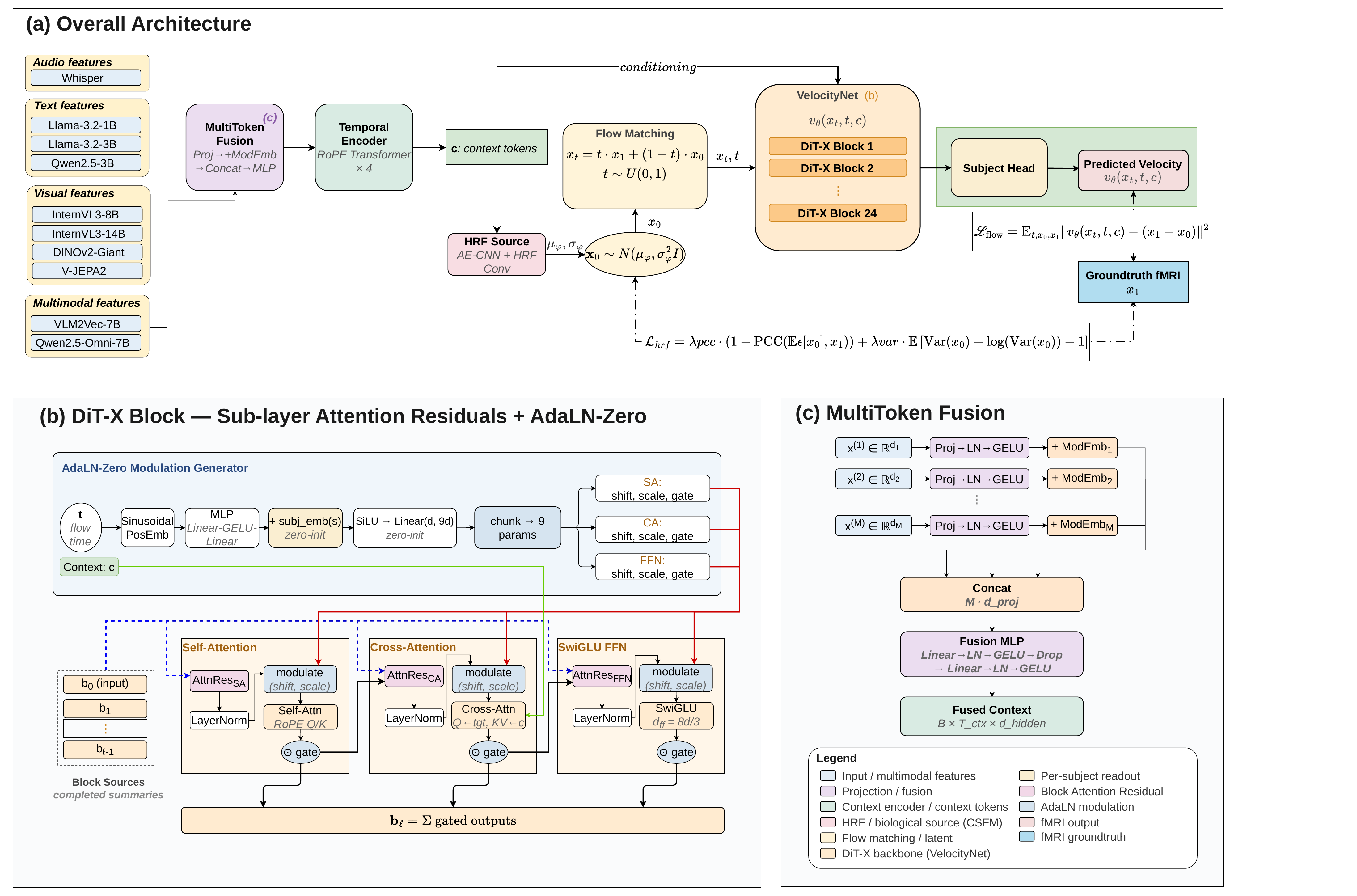}
\caption{\textbf{MultiFlow overview.} Ten frozen foundation backbones feed a multi-token fusion module and a RoPE temporal encoder. The encoded context conditions both the SISFM source generator (producing a stimulus-aligned starting distribution) and the DiT-X velocity decoder. Per-subject Yeo-network heads project the latent state to a 1000-parcel BOLD prediction.}
\label{fig:arch}
\end{figure*}

\subsection{Heterogeneous Multimodal Fusion}
\label{sec:fusion}

\noindent\textbf{Multimodal fusion.}
The ten backbones span nearly an order of magnitude in dimensionality ($d=1{,}280$ to $6{,}144$) (Figure~\ref{fig:arch}). Each TR-aligned stream $\mathbf{f}^{(i)}$ is projected to a common dimension $p=512$, summed with a learnable modality embedding $\mathbf{e}^{(i)} \in \mathbb{R}^{512}$, and concatenated along the feature axis. A two-layer MLP ($5120\!\to\!2048\!\to\!1024$) then produces the fused context $\mathbf{c}\!\in\!\mathbb{R}^{T_\text{ctx}\times 1024}$. Unlike the mean-pooled fusion in TRIBE~\cite{TRIBE}, this preserves per-modality information for later learned mixing.

\noindent\textbf{Temporal encoder.} 
A four-layer pre-norm Transformer with RoPE~\cite{RoPE} encodes relative stimulus-to-BOLD lag rather than absolute clip position. It runs bidirectionally over $T_\text{ctx}=101$ TRs and its output is centre-sliced to $T_\text{tgt}=50$ TRs, yielding $\mathbf{c}_\text{enc}\!\in\!\mathbb{R}^{50\times 1024}$ shared by the source generator and velocity decoder.

\subsection{Stimulus-Informed Source Flow Matching}
\label{sec:sisfm}

\begin{figure}[t]
\centering
\includegraphics[width=0.65\linewidth]{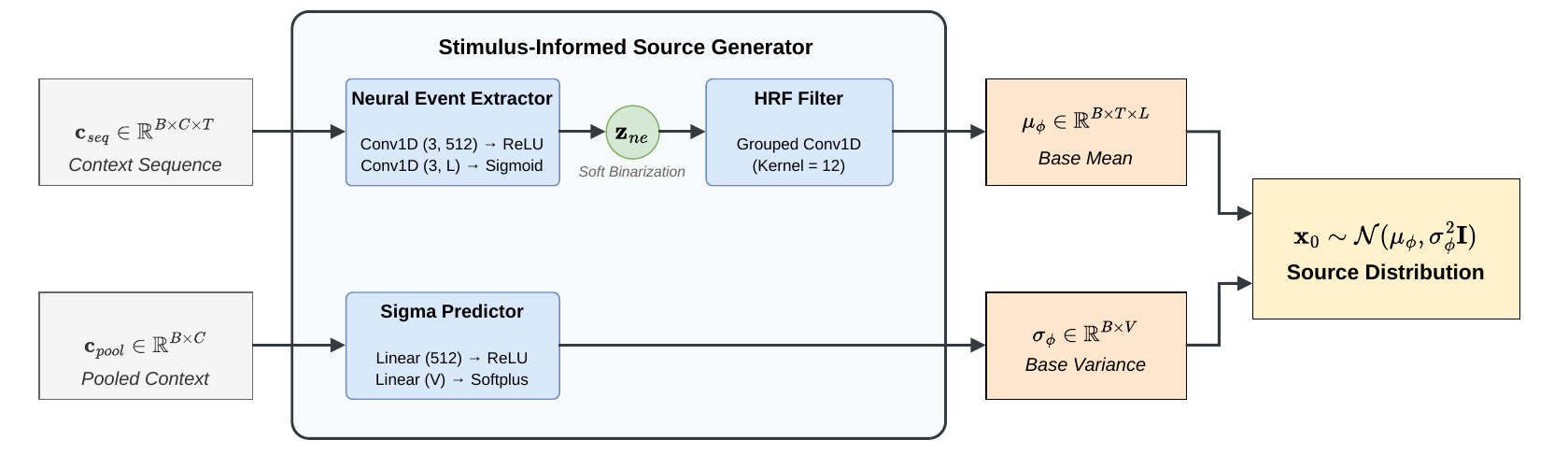}
\caption{\textbf{SISFM source generator.} A Conv1D--ReLU--Conv1D--Sigmoid stage produces a non-negative driver, convolved by a depthwise Conv1D with HRF-length kernel to yield $\boldsymbol{\mu}_\phi$. A pooled MLP--Softplus head produces $\sigma_\phi$.}
\label{fig:sisfm}
\end{figure}

\noindent\textbf{Source parameterisation.}
The stimulus-conditioned source distribution $p_0(\mathbf{z}_0\mid\mathbf{x})$ in Eq.~\eqref{eq:sisfm-source} is parameterised by a temporal operator whose receptive field matches the canonical HRF support window (Figure~\ref{fig:sisfm}). The mean $\boldsymbol{\mu}_\phi$ uses a Conv1D cascade over $\mathbf{c}_\text{enc}$ to produce a non-negative driver, convolved by a depthwise Conv1D whose kernel is unconstrained and whose length matches the canonical HRF support window; non-negativity applies to the driver, not the kernel, so the operator imposes an HRF \emph{support length} rather than an HRF \emph{shape}. The latent mean $\boldsymbol{\mu}_\phi^\text{lat}\!\in\!\mathbb{R}^{50\times 1024}$ is projected to voxel space by subject heads (Section~\ref{sec:heads}). The variance scale $\sigma_\phi(\mathbf{x}) = \mathrm{softplus}\bigl(h_\phi(\mathbf{x})\bigr) + 10^{-4}$ is predicted via a temporally pooled context and a two-layer MLP, yielding a per-voxel vector broadcast across TRs.

\noindent\textbf{Training objective.}
Joint training of $(\boldsymbol{\mu}_\phi,\sigma_\phi)$ and the velocity field $v_\theta$ is augmented to prevent the degenerate minimiser ($\boldsymbol{\mu}_\phi\!\to\!\mathbf{y}$, $\sigma_\phi\!\to\!0$) under which generative capacity collapses:
\begin{equation}
\mathcal{L}_\text{SISFM} = \mathcal{L}_\text{flow} + \lambda_\text{var}\,\mathcal{L}_\text{var} + \lambda_\text{pcc}\,\mathcal{L}_\text{pcc},
\label{eq:sisfm-loss}
\end{equation}
\begin{align}
\mathcal{L}_\text{var} = \mathbb{E}\bigl[\sigma_\phi^2 - \log\sigma_\phi^2 - 1\bigr],\qquad
\mathcal{L}_\text{pcc} = \mathbb{E}\bigl[1 - \mathrm{PCC}\bigl(\boldsymbol{\mu}_\phi(\mathbf{x}),\,\mathbf{y}\bigr)\bigr].
\end{align}
$\mathcal{L}_\text{var}$ is proportional to the KL divergence between $\mathcal{N}(0,\sigma_\phi^2)$ and a unit Gaussian, penalising variance collapse by pulling $\sigma_\phi$ toward unity. $\mathcal{L}_\text{pcc}$ provides scale-invariant supervision: maximising Pearson correlation pushes $\boldsymbol{\mu}_\phi$ to track the \emph{shape} of $\mathbf{y}$ without competing with $v_\theta$ for amplitude. From a validation grid search we fix $\lambda_\text{var}=0.05$ and $\lambda_\text{pcc}=1.0$; this soft balancing avoids the hard gradient-stopping used in~\cite{TRIBE,VIBE}. During training the mean voxel variance $\bar{\sigma}_\phi^2$ converges to $\approx 0.19$ while the source residual stays well above prediction error, indicating that the source does not collapse.

\subsection{DiT-X Decoder with Attention Residuals}
\label{sec:ditx}

While the source distribution provides a stimulus-aligned starting estimate, the velocity field transforms this state into the target BOLD response. This velocity field is parameterised with a 24-block DiT-X decoder~\cite{ManiFlow} (1024-dim, 16 heads). Each block combines self-attention over target tokens, cross-attention to $\mathbf{c}_\text{enc}$ (both with RoPE), and a SwiGLU FFN~\cite{Shazeer2020}, modulated by AdaLN-Zero~\cite{Peebles2023} with zero-initialised gates.

\noindent\textbf{Sub-layer Attention Residuals.} 
Following~\cite{AttentionResiduals}, the identity skip is replaced with learned softmax-weighted aggregation over all previous block summaries,
\begin{equation}
\mathrm{AR}\bigl(\{\mathbf{s}_n\}_{n=0}^{N}\bigr) = \sum_{n=0}^{N}\alpha_n \mathbf{s}_n,\qquad
\boldsymbol{\alpha} = \mathrm{softmax}\bigl(\mathbf{q}^{\!\top}\mathrm{RMSNorm}(\mathbf{s}_n)\bigr),
\end{equation}
where $\mathbf{q}\!\in\!\mathbb{R}^{1024}$ is a zero-initialised pseudo-query and summaries include the input embedding and all preceding block outputs. The operator is inserted at all three sub-layer positions, letting each velocity component draw on the depth where its relevant feature is most legible.

\subsection{Per-Subject Yeo-Network Heads}
\label{sec:heads}

The DiT-X output $\mathbf{h}\!\in\!\mathbb{R}^{50\times1024}$ is mapped to fMRI space by per-subject Yeo-network heads. For each of the seven Yeo networks~\cite{Yeo2011} (Visual, SomMot, DorsAttn, SalVentAttn, Limbic, Cont, Default) the head holds a weight tensor $W^{(k)}\!\in\!\mathbb{R}^{N_s\times1024\times 2n_k}$, with $N_s=4$ subjects and $2n_k$ the bilateral parcel count for network $k$ (so $\sum_k 2n_k = V=1000$). At inference the relevant subject's slice is applied to $\mathbf{h}$ and the seven outputs are concatenated to give $\hat{\mathbf{y}}\!\in\!\mathbb{R}^{50\times 1000}$. This lets functionally distinct cortical systems specialise their readout while sharing the upstream representation, a finer grain of subject conditioning than the single shared linear head used in TRIBE~\cite{TRIBE}.

\subsection{Inference}
\label{sec:inference}

At test time, $K$ independent samples $\mathbf{z}_0^{(k)}\!\sim\!p_0(\cdot\mid\mathbf{x})$ are drawn from the SISFM source (with temperature $\tau\!=\!0.2$ scaling $\sigma_\phi$) and the probability-flow ODE is integrated for each through the \emph{same} learned velocity field with a midpoint solver for $N=30$ steps on a square-root time grid $t_i=\sqrt{s_i}$, which places finer steps near $t=1$ where the trajectory becomes sharper. The $K$ predictions are then averaged, and long clips use sliding-window inference with overlap averaging. Unless stated otherwise, all reported numbers use $K\!=\!5$, at which the sampling gain saturates (Tables~\ref{tab:indist} and~\ref{tab:ood}). We call this \emph{source-averaged inference}, to make explicit that the $K$ draws come from the SISFM source of a \emph{single} trained checkpoint, unlike baselines that average 20--1000 re-trained models.

\FloatBarrier
\section{Experiments}
\label{sec:exp}

\subsection{Setup}

We evaluate on the Algonauts 2025 / CNeuroMod dataset~\cite{Algonauts2025,CNeuroMod} using subjects \texttt{sub-01}, \texttt{02}, \texttt{03}, and \texttt{05} (\texttt{sub-04} is excluded due to incomplete data). Training uses Friends S1--5 and four feature movies; S6 is validation, S7 is the in-distribution (ID) test split, and six held-out movies are the out-of-distribution (OOD) test split. fMRI volumes are z-scored, and we report per-parcel Pearson correlation (PCC) over the Schaefer-1000 parcellation. Ten frozen V/A/L/AV backbones are pooled to TR resolution ($D\!=\!31{,}872$; per-backbone details in the supplementary material). Training runs on a single RTX 3090 GPU for 100 epochs with AdamW ($\mathrm{lr}\!=\!10^{-4}$, cosine decay, 5\% warm-up), bfloat16, effective batch size 512, and weight EMA (decay 0.999); windows use $T_\text{ctx}\!=\!101$, $T_\text{tgt}\!=\!50$, stride 10, and jitter ${\pm}2$\,TRs.

\noindent\textbf{Baselines and protocol.} 
We compare MultiFlow (one trained model) against: (i) published ensemble results of TRIBE~\cite{TRIBE} (1000 models), VIBE~\cite{VIBE} (20 models), and MedARC~\cite{MedARC} (50 models, OOD only); (ii) reproduced single-run baselines (one trained model) under matched features/splits; and (iii) \textit{Regression DiT-X}, a deterministic control. Main results and ablations use DiT-X-Large (24 blocks, 1024-dim, 16 heads) trained on all subjects; ablations report mean S6 PCC.

All single-run baselines are retrained from the authors' published code under our features, data, and splits, with their released hyperparameters. One caveat applies to the reproduced TRIBE: its mean-pooling fusion was designed for the three-backbone set it was built around, so under our ten-backbone protocol it pools a considerably more heterogeneous input, which likely explains why it scores lowest of the single-run baselines despite being the strongest published ensemble. We therefore do not treat it as an upper bound on that method, and read the single-run column as a comparison of fusion designs under \emph{our} protocol rather than a ranking of the original systems.

\subsection{Main Results}

\begin{table}[tb]
\centering
\small
\renewcommand{\arraystretch}{1.15}
\setlength{\tabcolsep}{4pt}
\begin{tabular}{l cc ccccc}
\toprule
Method & \shortstack{\#Trained\\models} & \shortstack{\#Source\\draws} & Sub-01 & Sub-02 & Sub-03 & Sub-05 & Mean \\
\midrule
Ridge~\cite{Algonauts2025}
  & -- & --
  & 0.2152 & 0.2031 & 0.2189 & 0.1759 & 0.2033 \\
\midrule
\multicolumn{8}{l}{\textit{Single-run, matched training compute (1 trained model, same features/data/splits)}} \\
TRIBE~\cite{TRIBE}            & 1 & -- & 0.2827 & 0.2689 & 0.2817 & 0.2344 & 0.2670 \\
VIBE~\cite{VIBE}             & 1 & -- & 0.3201 & 0.3042 & 0.3093 & 0.2595 & 0.2983 \\
MedARC~\cite{MedARC}           & 1 & -- & 0.3271 & 0.3097 & 0.3209 & 0.2642 & 0.3055 \\
Regression DiT-X (ours, det.) & 1 & -- & 0.2887 & 0.2734 & 0.2807 & 0.2331 & 0.2690 \\
\midrule
\multicolumn{8}{l}{\textit{Full ensemble baselines}} \\
TRIBE~\cite{TRIBE}            & 1000 & -- & 0.3419 & 0.3239 & 0.3346 & 0.2775 & 0.3195 \\
VIBE~\cite{VIBE}             & 20  & -- & \textbf{0.3503} & \textbf{0.3309} & \textbf{0.3410} & \textbf{0.2849} & \textbf{0.3268} \\
\midrule
\textbf{MultiFlow (ours)}     & 1 & 5 & 0.3398 & 0.3217 & 0.3275 & 0.2760 & 0.3163 \\
\bottomrule
\end{tabular}
\caption{\textbf{In-distribution performance (Friends S7).} Per-subject and mean per-parcel PCC ($\uparrow$) over Schaefer-1000. \emph{\#Trained models} counts independently trained checkpoints; \emph{\#Source draws} counts samples averaged from the SISFM source of a \emph{single} checkpoint (source-averaged inference, Section~\ref{sec:inference}), which adds no training cost. ``Matched training compute'' refers to the \emph{training} budget: all single-run rows share MultiFlow's features, data, splits, and frozen backbones, whose inference cost dominates. \textit{Regression DiT-X} is identical to MultiFlow except that the SISFM source and variance head are removed and MSE is used, so its gap to MultiFlow isolates the generative formulation. MultiFlow uses $K=5$ draws ($\tau=0.2$); bold is best.}
\label{tab:indist}
\end{table}

\begin{table}[tb]
\centering
\small
\renewcommand{\arraystretch}{1.15}
\setlength{\tabcolsep}{4pt}
\begin{tabular}{l cc ccccc}
\toprule
Method & \shortstack{\#Trained\\models} & \shortstack{\#Source\\draws} & Sub-01 & Sub-02 & Sub-03 & Sub-05 & Mean \\
\midrule
Ridge~\cite{Algonauts2025}
  & -- & -- & 0.0986 & 0.0859 & 0.1021 & 0.0715 & 0.0895 \\
\midrule
\multicolumn{8}{l}{\textit{Single-run, matched training compute (1 trained model, same features/data/splits)}} \\
TRIBE~\cite{TRIBE}            & 1 & -- & 0.1875 & 0.1726 & 0.1838 & 0.1381 & 0.1705 \\
VIBE~\cite{VIBE}             & 1 & -- & 0.2172 & 0.1921 & 0.2135 & 0.1646 & 0.1968 \\
MedARC~\cite{MedARC}           & 1 & -- & 0.2282 & 0.1958 & 0.2275 & 0.1702 & 0.2054 \\
Regression DiT-X (ours, det.) & 1 & -- & 0.1372 & 0.1291 & 0.1463 & 0.1147 & 0.1318 \\
\midrule
\multicolumn{8}{l}{\textit{Full ensemble baselines}} \\
TRIBE~\cite{TRIBE}            & 1000 & -- & 0.2381 & \textbf{0.2105} & 0.2377 & 0.1720 & 0.2146 \\
VIBE~\cite{VIBE}             & 20  & -- & 0.2392 & 0.2080 & 0.2294 & 0.1721 & 0.2122 \\
MedARC~\cite{MedARC}           & 50  & -- & \textbf{0.2458} & 0.2071 & \textbf{0.2384} & \textbf{0.1772} & \textbf{0.2171} \\
\midrule
\textbf{MultiFlow (ours)}     & 1 & 5 & 0.2319 & 0.2033 & 0.2218 & 0.1752 & 0.2080 \\
\bottomrule
\end{tabular}
\caption{\textbf{Out-of-distribution performance (6 held-out movies).} Per-subject and mean per-parcel PCC ($\uparrow$) over Schaefer-1000. Columns as in Table~\ref{tab:indist}: \emph{\#Trained models} counts independently trained checkpoints, \emph{\#Source draws} counts averaged samples from a single checkpoint. \textit{Regression DiT-X} is the architecture-matched MSE control (see Table~\ref{tab:indist}). MultiFlow uses $K=5$ draws ($\tau=0.2$). Bold denotes the best result.}
\label{tab:ood}
\end{table}

\noindent\textbf{Analysis of main results.}
Results are analysed in-distribution (ID, Table~\ref{tab:indist}) and
out-of-distribution (OOD, Table~\ref{tab:ood}). Two controls attribute any
improvement to the right cause: a \emph{matched-training-compute} control (all
single-run rows share MultiFlow's features, data, and splits) and a
\emph{matched-architecture} control (\textit{Regression DiT-X}, which removes the
SISFM source but keeps the backbone fixed and is trained with MSE).

\textit{System level (matched compute).} Among single-run models, MultiFlow
attains the highest mean PCC on both splits \emph{at $K{=}5$ source draws}. Specifically, on ID, it scores $0.3163$ vs.\ $0.2670$ (TRIBE),
$0.2983$ (VIBE), and $0.3055$ (MedARC); on OOD, it scores $0.2080$, above all three single-run baselines (TRIBE $0.1705$, VIBE $0.1968$, MedARC $0.2054$). At a single draw ($K{=}1$) the margins narrow and the OOD ordering no longer holds ($0.1983$ against MedARC's $0.2054$). This
makes the full system strongest at $1\times$ training compute, but because these baselines differ in \emph{both} architecture and objective, it does not by itself attribute the gain to the generative formulation.

\textit{Formulation level (matched architecture).} Holding the DiT-X backbone
fixed and replacing MSE with SISFM lifts mean PCC from $0.2690$ to $0.3163$ on ID
and from $0.1318$ to $0.2080$ on OOD (\textit{Regression DiT-X} vs.\ MultiFlow,
Tables~\ref{tab:indist}--\ref{tab:ood}), isolating the contribution of the
generative formulation on the held-out test splits. The architecture alone does not
exceed the strongest single-run baseline: Regression DiT-X ($0.2690$) falls
$0.0365$ below MedARC ($0.3055$) on ID, and SISFM closes that deficit and adds
$+0.0108$, giving the full $+0.0473$ gain over the architecture-matched control. On
OOD the effect is larger: SISFM contributes $+0.0762$, lifting the architecture from
$0.1318$, below the $0.1705$--$0.2054$ range of the single-run baselines, to $0.2080$
above all of them; Section~\ref{sec:uq} attributes this to an overfitting
regularisation effect. This reproduces on S7/OOD the same-architecture trend seen on
S6 in Table~\ref{tab:gen} (MSE $0.2760\!\to\!$ CFM $0.2973\!\to\!$ SISFM $0.3213$).

\textit{Comparison with ensembles.} Against published ensemble results,
single-checkpoint MultiFlow trails the 1000-model TRIBE ensemble on ID by only $0.003$ PCC ($0.3163$ vs.\ $0.3195$), trailing the strongest ID ensemble (VIBE-20) by $0.010$ PCC. On OOD, MultiFlow trails the strongest ensemble (MedARC-50) by $0.009$ PCC ($0.004$--$0.009$ across ensembles), while on Sub-05 it approaches the best ensemble ($0.1752$ vs.\ MedARC-50's $0.1772$). We do not claim parity
with ensembled SOTA on accuracy; rather, MultiFlow recovers most of the
ensembling benefit using a single trained model rather than an ensemble of $20$--$1000$ independently retrained
checkpoints. A significance test on these margins is not possible: S7 and the OOD movies are
scored by the provider's private server, which returns only an aggregate PCC, and
the published ensembles release no per-parcel outputs. We therefore report the
ensemble margins descriptively and disclaim parity; Section~\ref{sec:uq} gives
per-parcel statistics for the testable comparisons, those against our own
architecture-matched control.

\textit{Source-averaging gain.} Drawing $K=5$ samples from one checkpoint
improves over a single sample ($K=1$): $0.3057\!\to\!0.3163$ on ID and
$0.1983\!\to\!0.2080$ on OOD, after which the gain saturates.

\textit{Generative capabilities.} Beyond accuracy, sampling from the
SISFM source yields samples from the predictive distribution and a
stimulus-conditioned scale $\sigma_\phi(\mathbf{x})$ from the same checkpoint, which are capabilities
a single deterministic regressor does not provide.

\FloatBarrier
\subsection{Cross-Subject Generalisation}

Generalisation to unseen subjects probes how much of a model's fit is carried by subject-specific readout parameters rather than by the shared upstream representation. Since TRIBE, VIBE, and MedARC do not report cross-subject results on this dataset, evaluation is performed using reproduced single-run baselines (same features, pipeline, no ensembling). This ensures a matched-training-compute comparison that isolates algorithmic differences from ensembling effects. Ensembled cross-subject numbers are unavailable for comparison: none are published, and producing them would require retraining each baseline's full $20$--$1000$-model ensemble once per leave-one-subject-out fold, so Table~\ref{tab:cross} is a single-run comparison by necessity. Table~\ref{tab:cross} reports leave-one-subject-out PCC: models trained on three subjects, evaluated on the held-out fourth, averaged over three clips (Friends S6, \textit{Figures}, \textit{Life}). For the held-out subject, inference uses the mean of the trained subjects' Yeo-network heads as the readout projection, so no subject-specific parameters are available for the test subject.

\begin{table}[!htbp]
\centering
\small
\setlength{\tabcolsep}{4pt}
\renewcommand{\arraystretch}{1.15}
\begin{tabular}{l cccc}
\toprule
 & \multicolumn{4}{c}{Hold-out (test) subject (model trained on remaining 3)} \\
\cmidrule(lr){2-5}
Method
  & \shortstack{Sub-01 \\ {\scriptsize train: 02, 03, 05}}
  & \shortstack{Sub-02 \\ {\scriptsize train: 01, 03, 05}}
  & \shortstack{Sub-03 \\ {\scriptsize train: 01, 02, 05}}
  & \shortstack{Sub-05 \\ {\scriptsize train: 01, 02, 03}} \\
\midrule
TRIBE~\cite{TRIBE}      & 0.1625 & 0.1584 & 0.1691 & 0.1242 \\
VIBE~\cite{VIBE}      & 0.2376 & 0.2442 & 0.2499 & 0.2336 \\
MedARC~\cite{MedARC}      & 0.2007 & 0.2040 & 0.2232 & 0.1855 \\
\midrule
\textbf{MultiFlow (ours)} & \textbf{0.2388} & \textbf{0.2507} & \textbf{0.2502} & \textbf{0.2442} \\
\bottomrule
\end{tabular}
\caption{\textbf{Cross-subject generalisation (leave-one-subject-out).} Column indicates the held-out test subject. PCC is averaged over three clips (Friends S6, \textit{Figures}, \textit{Life}). Baselines are single-run reproductions (no ensembling, same features/pipeline as MultiFlow). Bold denotes the best result.}
\label{tab:cross}
\end{table}

\noindent\textbf{Analysis.}
Under matched single-run conditions, MultiFlow matches or exceeds the strongest single-run baseline (VIBE~\cite{VIBE}) on every held-out subject, with the largest margin on Sub-05 ($0.2442$ vs.\ $0.2336$). We attribute this to two factors: the generative formulation regularises implicitly (Section~\ref{sec:uq}), reducing overfitting to training subjects, and the per-subject Yeo-network heads share an upstream representation, which eases transfer when they are averaged for an unseen subject. The absolute numbers set the interpretation: leave-one-subject-out PCC spans $0.239$--$0.251$ against $0.3163$ in-distribution, a shortfall of roughly a quarter of the ID score. Table~\ref{tab:cross} should therefore be read as evidence of \emph{relative robustness} under a matched protocol, meaning MultiFlow degrades less than the baselines when the test subject is unseen, and not as evidence of a subject-invariant representation. Establishing the latter would require broader subject coverage and readouts that are meta-learned rather than averaged (Section~\ref{sec:conclusion}).

\subsection{Spatial Mapping and Qualitative Visualisation}
\label{sec:qualitative}

We project per-parcel S6 PCC onto the Schaefer-1000 inflated cortical surface and, for six representative regions, plot region-averaged predicted BOLD against ground truth over a full episode.

\begin{figure*}[t]
\centering
\includegraphics[width=0.52\linewidth]{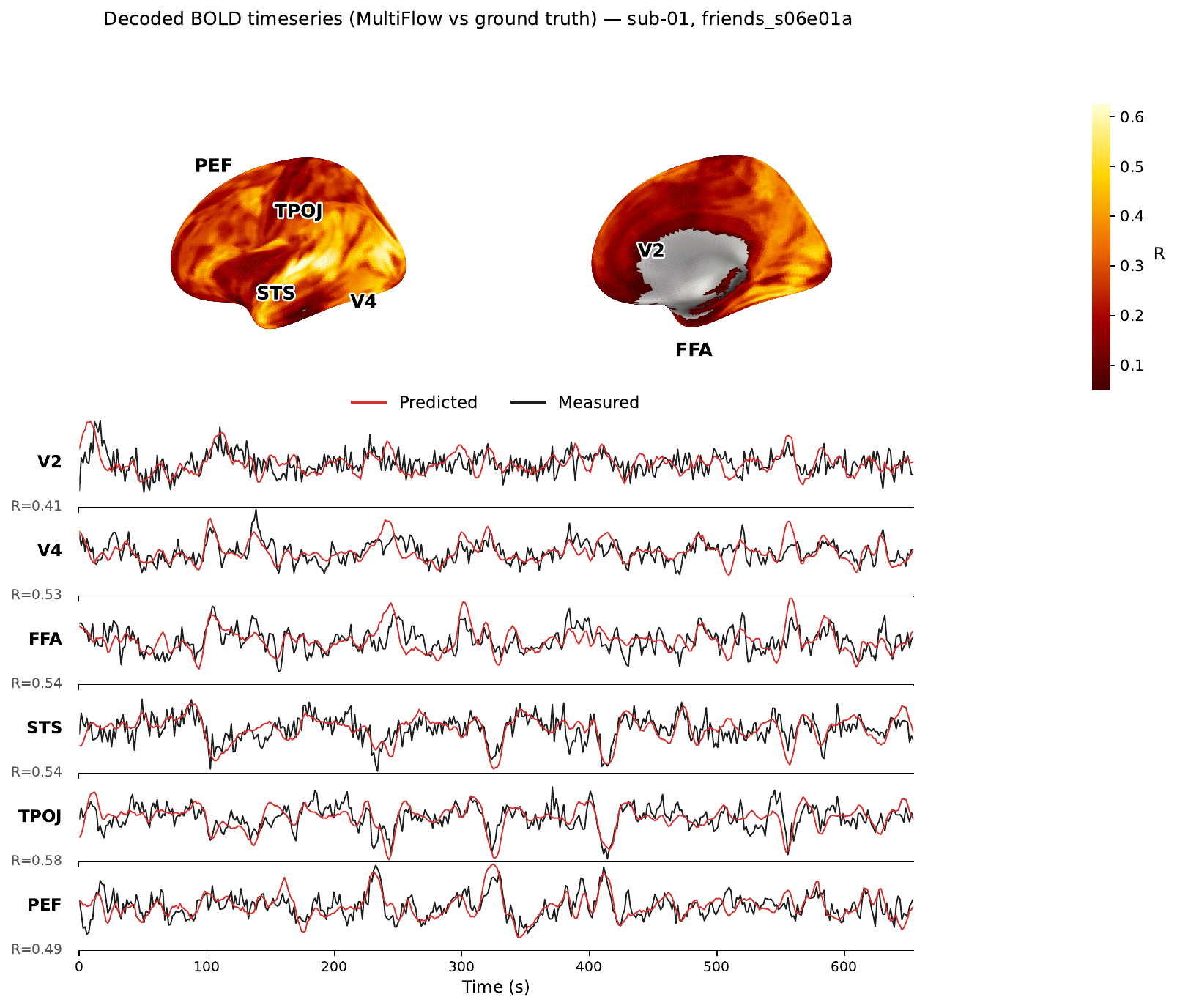}
\caption{\textbf{Predicted BOLD timeseries across the cortical hierarchy (sub-01, Friends s06e01a).} \emph{Top:} per-parcel PCC on the fsaverage5 inflated surface (LH lateral, RH medial), six regions annotated. \emph{Bottom:} predicted (red) vs.\ measured (black) BOLD averaged over each region's parcels across $\sim$656\,s; parcel-mean R on the full S6 holdout is given next to each track.}
\label{fig:s6_region_timeseries}
\end{figure*}

\noindent\textbf{Cortical hierarchy and temporal fidelity.} 
Figure~\ref{fig:s6_region_timeseries} (top) shows the classical encoding gradient~\cite{Yamins2016}: per-parcel PCC peaks in ventral and lateral occipito-temporal cortex (V4, fusiform, STS, TPOJ), decays anteriorly, and varies smoothly across the surface. The traces (bottom) make the hierarchy concrete: TPOJ ($R{=}0.58$), STS / FFA ($R{=}0.54$) and V4 ($R{=}0.53$) sit in the upper achievable range, while PEF ($R{=}0.49$) and the small-n medial V2 cluster ($R{=}0.41$) drop slightly but stay temporally aligned. For $R{\geq}0.5$ regions the prediction tracks BOLD peaks and troughs at the TR scale, including excursions above 1\,SD lasting 10--30\,s, consistent with the HRF-length source mean capturing much of the temporal structure before the velocity field acts.

\subsection{Component Ablations}

All ablations use the DiT-X-Large variant and are trained on all four subjects (\texttt{sub-01}, \texttt{02}, \texttt{03}, \texttt{05}) using Friends Seasons~1--5 plus the four feature movies, with Friends Season~6 held out for evaluation (mean PCC across the four subjects). S6 is the benchmark's designated validation split and its ground truth is available locally, enabling rapid iterative comparison across many ablation variants. The S7 and OOD labels are withheld by the dataset provider and can be scored only through its evaluation server, which returns a single aggregate PCC per submission; iterative ablation and per-parcel analysis are therefore not possible on those splits, and they are reserved for the final system-level evaluation in Tables~\ref{tab:indist}--\ref{tab:ood}.

\noindent\textbf{Generative versus deterministic encoding.} 
Table~\ref{tab:gen} compares three formulations of the same architecture: Regression DiT-X (MSE loss), CFM with a standard Gaussian source, and the full SISFM model.

\begin{table}[htbp]
\centering
\setlength{\tabcolsep}{3pt}
\begin{minipage}[t]{0.55\linewidth}
  \centering
  \footnotesize
  \begin{tabular}{lccc}
  \toprule
  Model & Source $p_0$ & Training loss & S6 PCC ($\uparrow$)\\
  \midrule
  Regression DiT-X & - & MSE & 0.2760 \\
  CFM + Gaussian & $\mathcal{N}(\mathbf{0},\mathbf{I})$ & $\mathcal{L}_\text{flow}$ & 0.2973 \\
  CFM + SISFM & $p_0(\cdot\!\mid\!\mathbf{x})$ & Eq.~\ref{eq:sisfm-loss} & \textbf{0.3213} \\
  \bottomrule
  \end{tabular}
  \caption{\textbf{Generative vs.\ deterministic encoding.} Same architecture, three formulations. Across three inference seeds the MSE$\to$CFM$\to$SISFM ordering is preserved, with a per-row range $\le 0.002$ PCC, well below the reported gaps.}
  \label{tab:gen}

  \vspace{5pt}

  \begin{tabular}{lc}
  \toprule
  Decoder & S6 PCC ($\uparrow$) \\
  \midrule
  MLP (3-layer) & 0.2565 \\
  UNet  & 0.2592 \\
  DiT & 0.3035 \\
  DiT-X + Attention Residuals & \textbf{0.3213} \\
  \bottomrule
  \end{tabular}
  \caption{\textbf{Decoder backbone.} Effect of cross-attention and sub-layer Attention Residuals.}
  \label{tab:dec}
\end{minipage}
\hfill
\begin{minipage}[t]{0.43\linewidth}
  \centering
  \footnotesize
  \begin{tabular}{lccccl}
  \toprule
  Setting & V & A & L & AV & S6 PCC $\uparrow$ \\
  \midrule
  V only             & \checkmark & -          & -          & -          & 0.2043 \\
  A only             & -          & \checkmark & -          & -          & 0.1597 \\
  L only             & -          & -          & \checkmark & -          & 0.2170 \\
  V + A              & \checkmark & \checkmark & -          & -          & 0.2821 \\
  V + L              & \checkmark & -          & \checkmark & -          & 0.2995 \\
  A + L              & -          & \checkmark & \checkmark & -          & 0.2872 \\
  V + A + L          & \checkmark & \checkmark & \checkmark & -          & 0.3020 \\
  All                & \checkmark & \checkmark & \checkmark & \checkmark & \textbf{0.3213} \\
  \bottomrule
  \end{tabular}
  \caption{\textbf{Modality coverage.} Factorial ablation over the V, A, L, and AV groups: single-modality rows give standalone predictive power, pairwise rows show cross-modal complementarity, and the last two rows give the marginal gain of the full stack.}
  \label{tab:mod}
\end{minipage}
\end{table}

Replacing MSE regression with standard-Gaussian CFM improves PCC by $+0.0213$, indicating that a generative formulation benefits encoding even without a stimulus-conditioned source. SISFM adds a further $+0.0240$, the largest single gain: by learning an HRF-length source mean $\boldsymbol{\mu}_\phi$ and variance $\sigma_\phi$, it offloads basic hemodynamics from the velocity field, which then models only residual variability. The total gain over regression is $+0.0453$ ($+16.4\%$ relative).

\noindent\textbf{Decoder backbone.} 
Table~\ref{tab:dec} compares the DiT-X decoder against an MLP, a UNet, and 1D-DiT without cross-attention.

MLP and UNet decoders perform comparably low ($0.2565$ vs.\ $0.2592$): MLPs ignore temporal structure, and UNets are constrained by the local receptive field of 1D convolutions. Replacing the UNet with a self-attentive DiT adds $+0.0443$, indicating that BOLD at any TR depends on long-range temporal context. Adding cross-attention and sub-layer Attention Residuals (DiT-X + AR) contributes a further $+0.0178$: cross-attention in every block lets the decoder query the stimulus context at different temporal lags, and Attention Residuals allow depth-selective routing between low-level sensory and semantic features.

\noindent\textbf{Modality coverage.} 
Table~\ref{tab:mod} isolates the contribution of each modality group through a factorial ablation.

Fusing several backbones already helps within a modality: the strongest \emph{single} visual backbone (InternVL3-14B, $0.1932$) trails the fused V-only setting ($0.2043$), so part of the gain is within-modality complementarity rather than cross-modal binding. Language alone (0.217) slightly outperforms vision alone (0.204), consistent with the recruitment of semantic and social networks during narrative viewing~\cite{Huth2016}. Pairwise combinations exceed their constituents, V+A (0.282) over V-only (0.204) and A-only (0.159), and adding jointly pre-trained audio-visual backbones to V+A+L gives the largest marginal gain, from 0.3020 to 0.3213, consistent with joint pre-training capturing cross-modal binding.

\subsection{Generative Analysis: Robustness, UQ, and Diversity}
\label{sec:uq}

This section evaluates step robustness, uncertainty quantification (UQ), and sample diversity on the validation split (Friends S6).

\begin{figure}[t]
\centering
\begin{minipage}[t]{0.48\linewidth}
\centering
\includegraphics[width=\linewidth]{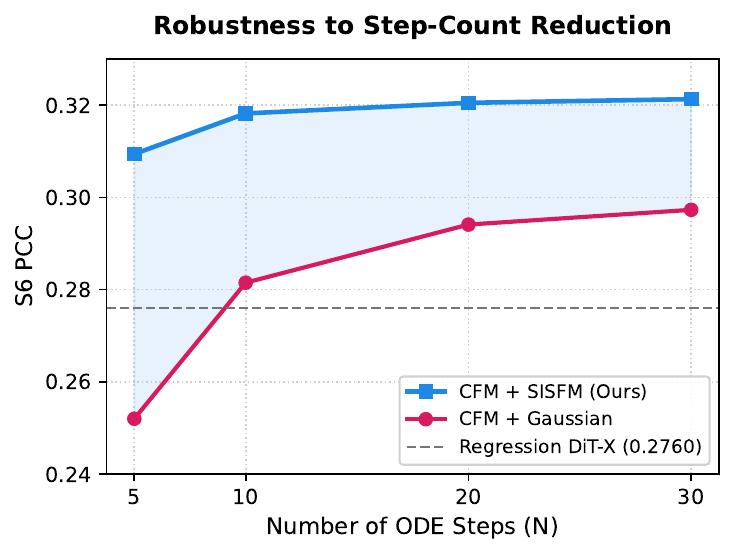}
\caption{\textbf{Step-count robustness (S6 PCC).} From $N{=}30$ to $N{=}5$ steps, SISFM (blue) retains accuracy while standard CFM (red) decays, falling below the Regression DiT-X baseline (dashed grey) at low $N$.}
\label{fig:nfe_robustness}
\end{minipage}
\hfill
\begin{minipage}[t]{0.48\linewidth}
\centering
\includegraphics[width=\linewidth]{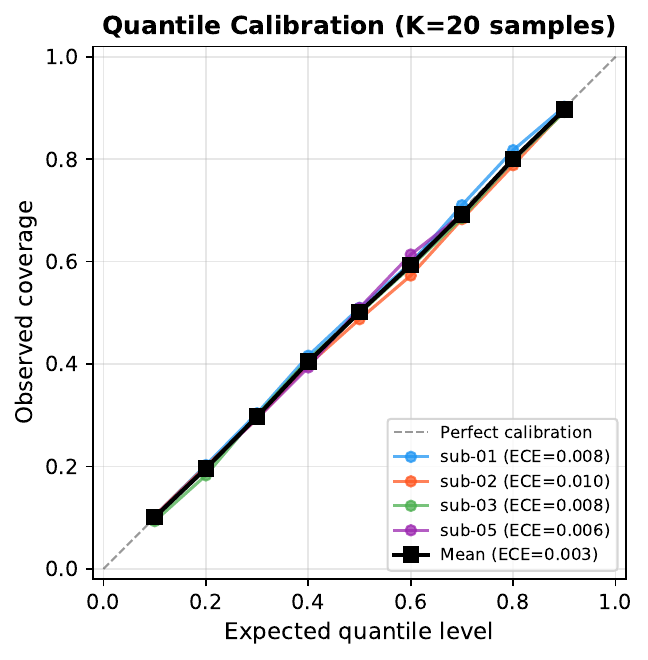}
\caption{\textbf{Quantile calibration (S6, $K{=}20$).} Observed coverage vs.\ expected quantile level; proximity to the diagonal indicates calibrated predictive intervals from a single checkpoint. Per-subject expected calibration error ranges over $0.006$--$0.010$.}
\label{fig:calib}
\end{minipage}
\end{figure}

\noindent\textbf{Path length and step-count robustness.}
Straightness of the learned velocity field is governed by the source--data coupling, whose expected squared displacement is $\mathbb{E}_{\boldsymbol{\epsilon}}\,\|\mathbf{y}-\mathbf{z}_0\|^2$ (with $\mathbf{z}_0=\boldsymbol{\mu}_\phi+\sigma_\phi\boldsymbol{\epsilon}$). This equals $\|\mathbf{y}\|^2+d$ for a Gaussian source and $\|\mathbf{y}-\boldsymbol{\mu}_\phi\|^2+d\,\bar{\sigma}_\phi^2$ for SISFM ($d=50{,}000$, with $\bar{\sigma}_\phi^2\!\approx\!0.19$ the converged training-time variance). On S6 the expected length drops from $98.3$k to $57.4$k ($-41.6\%$), almost entirely through variance shrinkage: the noise term $d\bar{\sigma}_\phi^2$ contracts from $50$k to $9.5$k, while the mean term $\|\mathbf{y}-\boldsymbol{\mu}_\phi\|^2$ falls by less than $1\%$ (from $\sim$$48.3$k to $\sim$$47.9$k), because the scale-invariant objective $\mathcal{L}_\text{pcc}$ makes $\boldsymbol{\mu}_\phi$ a shape-aligned prior (PCC $0.26$) rather than an amplitude-matched one. The tighter coupling yields straighter trajectories, and at inference samples are drawn closer still to $\boldsymbol{\mu}_\phi$ (temperature $\tau{=}0.2$ scaling $\sigma_\phi$), so the probability-flow ODE integrates accurately under few steps.
As shown in Table~\ref{tab:nfe} and Figure~\ref{fig:nfe_robustness}, reducing ODE steps from $N{=}30$ to $N{=}5$ collapses standard CFM to $0.2520$ while SISFM holds $0.3094$, widening $\Delta$ from $+0.0240$ to $+0.0574$ and confirming that a shorter transport path buys step-efficiency at test time.

\begin{table}[htbp]
\centering\footnotesize
\begin{tabular}{lcccc}
\toprule
ODE steps $N$ & 5 & 10 & 20 & 30 \\
\midrule
CFM + Gaussian & 0.2520 & 0.2815 & 0.2941 & 0.2973 \\
CFM + SISFM    & 0.3094 & 0.3182 & 0.3205 & 0.3213 \\
$\Delta$ (SISFM$-$Gauss) & $+$0.0574 & $+$0.0367 & $+$0.0264 & $+$0.0240 \\
\bottomrule
\end{tabular}
\caption{\textbf{Step-count robustness (S6 PCC, $\uparrow$).} The SISFM source preserves accuracy under step-count reduction better than a Gaussian source, the behaviour predicted by its shorter expected transport path.}
\label{tab:nfe}
\end{table}

\noindent\textbf{Source averaging and regularisation.}
At test time, drawing $K=5$ independent samples $\mathbf{z}_0^{(k)}\!\sim\!p_0$ and integrating them through the same velocity field (source-averaged inference) improves performance from $0.3057$ ($K=1$) to $0.3163$ ($K=5$) on ID, recovering the variance reduction of deep ensembles without retraining (Tables~\ref{tab:indist}--\ref{tab:ood}). The generative formulation also acts as a regulariser: under MSE with identical hyperparameters, Regression DiT-X drops $51.0\%$ in mean PCC from ID to OOD ($0.2690 \to 0.1318$), whereas MultiFlow drops $34.2\%$ ($0.3163 \to 0.2080$), comparable to the $32$--$34\%$ of VIBE ($0.2983 \to 0.1968$) and MedARC ($0.3055 \to 0.2054$) while remaining more accurate on both splits.

\noindent\textbf{Uncertainty quantification and calibration.}
We run inference with $K=20$ samples on S6. First, the learned per-voxel predictive scale $\sigma_\phi(\mathbf{x})\!\in\!\mathbb{R}^V$ correlates with per-voxel prediction error (Spearman $\rho=0.36 \pm 0.04$, $p<10^{-5}$), so the source variance is a usable confidence signal. Second, treating the spread of the $K=20$ samples as a predictive distribution, the expected calibration error, the mean absolute gap between observed coverage and expected quantile level over 9 levels ($p\in\{0.1,\dots,0.9\}$), ranges from $0.006$ to $0.010$ across the four subjects (Figure~\ref{fig:calib}). Third, the sample standard deviation does not collapse (mean $0.32 \pm 0.03$) and is higher where error is larger (Spearman $\rho=0.42$, $p<10^{-6}$), so MultiFlow expresses heteroscedastic uncertainty rather than degenerate point estimates.

\noindent\textbf{Scope and biological interpretability.}
Three qualifications bound these results. First, the calibration result is a property
of the \emph{transported sample distribution}, not of the magnitude of $\sigma_\phi$:
the source scale is rescaled at inference by $\tau{=}0.2$, so calibration is produced
by the source and the velocity field jointly, and $\sigma_\phi$ alone carries no
calibrated interval. Second, $\sigma_\phi$ is stimulus-conditioned \emph{by
construction}, being a function of $\mathbf{x}$; what we demonstrate is that its
spatial profile tracks per-voxel error, not that it varies with stimulus
\emph{content}. Third, our experiments cannot say how much of the modelled
variability is genuine neurophysiological noise and how much is epistemic,
reflecting what the model has not learned from limited data.
That separation requires an observable this dataset lacks under a held-out protocol:
repeated presentations of the same stimulus to the same participant, or a parameter
ensemble whose disagreement isolates the epistemic component. We therefore retain two
readings of the SISFM source, as an implicit regulariser (the ID$\to$OOD degradation
above) and as an empirically calibrated predictive distribution, and retire a third: we do
not claim $\sigma_\phi$ measures biological response variability. A
repeat-presentation protocol, or a parameter ensemble evaluated against measured
test--retest variance, would settle this.

\section{Conclusion}
\label{sec:conclusion}

\noindent\textbf{Limitations and future work.}
\textbf{(i)}~$\sigma_\phi$ is a \emph{predictive} scale, and whether it decomposes
into physiological and cognitive components is untested here: the dataset offers no
repeated presentations under a held-out protocol, so $\sigma_\phi$ is validated
against prediction error rather than measured response variability
(Section~\ref{sec:uq}). It is also per-voxel but shared across TRs within a window;
a graph-structured or per-TR covariance could capture spatiotemporal
heteroscedasticity, and lightweight adaptation of the frozen backbones may benefit
the Default Mode and Control networks.
\textbf{(ii)}~Cross-subject results establish relative robustness, not
subject-invariance: leave-one-subject-out PCC ($0.239$--$0.251$) sits well below
in-distribution accuracy ($0.3163$), and transfer still relies on per-subject heads
and on SISFM hyperparameters selected on \texttt{sub-01}. Broader subject coverage
and meta-learned readouts are the route forward.
\textbf{(iii)}~Inference cost constrains online use: a midpoint solver at $N{=}30$
costs roughly $60$ network evaluations per source draw, short of real-time BCI
budgets. The shortened transport path buys headroom, since at
$N{=}5$ SISFM still reaches $0.3094$ against $0.2760$ for the deterministic control
and $0.2520$ for Gaussian CFM (Table~\ref{tab:nfe}), and distillation would extend
this further.

\noindent\textbf{Summary.}
SISFM conditions the flow-matching source on the stimulus; its instantiation MultiFlow reaches $0.3163$ ID and $0.2080$ OOD PCC from one checkpoint, within $0.003$ and $0.009$ PCC of 20--1000-model ensembles, and on a fixed architecture lifts S6 PCC from $0.2760$ to $0.3213$ (Table~\ref{tab:gen}) while also providing a per-voxel predictive scale and calibrated sample spread (Section~\ref{sec:uq}). The principle extends wherever a fixed Gaussian source ignores information available at inference, as in speech synthesis~\cite{Voicebox}, molecular docking~\cite{DiffDock}, and molecular design~\cite{EVOSYNTH}.

\section*{Acknowledgements}
This work was supported by the Institute of Information \& Communications Technology Planning \& Evaluation (IITP) under the Artificial Intelligence Convergence Innovation Human Resources Development (IITP-2023-RS-2023-00256629) grant funded by the Korea government (MSIT), the ITRC (Information Technology Research Center) support program (IITP-2026-RS-2024-00437718) supervised by IITP.

\bibliography{egbib}
\end{document}